\def\year{2022}\relax
\documentclass[letterpaper]{article} % DO NOT CHANGE THIS
\usepackage{aaai22}  % DO NOT CHANGE THIS
\usepackage{times}  % DO NOT CHANGE THIS
\usepackage{helvet}  % DO NOT CHANGE THIS
\usepackage{courier}  % DO NOT CHANGE THIS
\usepackage[hyphens]{url}  % DO NOT CHANGE THIS
\usepackage{graphicx} % DO NOT CHANGE THIS
\usepackage{natbib}  % DO NOT CHANGE THIS AND DO NOT ADD ANY OPTIONS TO IT
\usepackage{caption} % DO NOT CHANGE THIS AND DO NOT ADD ANY OPTIONS TO IT
\DeclareCaptionStyle{ruled}{labelfont=normalfont,labelsep=colon,strut=off} % DO NOT CHANGE THIS
\usepackage{algorithm}
\usepackage{algorithmic}
\usepackage{amsmath}
\usepackage{amsthm}
\newtheorem{definition}{Definition}
\nocopyright
\title{Fairness-aware Naive Bayes Classifier for Data with Multiple Sensitive Features}
\author {
    Stelios Boulitsakis-Logothetis
}
\affiliations{
    University of Durham\\
    Durham, United Kingdom\\
    stelios.b.logothetis@gmail.com
}

\begin{document}

\maketitle

\begin{abstract}
Fairness-aware machine learning seeks to maximise utility in generating predictions while avoiding unfair discrimination based on sensitive attributes such as race, sex, religion, etc. An important line of work in this field is enforcing fairness during the training step of a classifier. A simple yet effective binary classification algorithm that follows this strategy is two-naive-Bayes (2NB), which enforces statistical parity - requiring that the groups comprising the dataset receive positive labels with the same likelihood. In this paper, we generalise this algorithm into N-naive-Bayes (NNB) to eliminate the simplification of assuming only two sensitive groups in the data and instead apply it to an arbitrary number of groups. 

We propose an extension of the original algorithm's statistical parity constraint and the post-processing routine that enforces statistical independence of the label and the single sensitive attribute. Then, we investigate its application on data with multiple sensitive features and propose a new constraint and post-processing routine to enforce \emph{differential fairness}, an extension of established group-fairness constraints focused on intersectionalities. We empirically demonstrate the effectiveness of the NNB algorithm on US Census datasets and compare its accuracy and debiasing performance, as measured by disparate impact and DF-$\epsilon$ score, with similar group-fairness algorithms. Finally, we lay out important considerations users should be aware of before incorporating this algorithm into their application, and direct them to further reading on the pros, cons, and ethical implications of using statistical parity as a fairness criterion. 
\end{abstract}

\section{Introduction}

Today, countless machine learning-based systems are in use that autonomously make decisions or aid human decision-makers in applications that significantly impact individuals' lives. This has made it vital to develop ways of ensuring these models are trustworthy, ethical, and fair. The field of fairness-aware machine learning is centered on enhancing the fairness, explainability, and auditability of ML models. A goal many research works in this field share is to maximise utility in generating predictions while avoiding discrimination against people based on specific sensitive attributes, such as race, sex, religion, nationality, etc.

Researchers have devised many formalisations to try and capture intuitive notions of fairness, each with different priorities and limitations. We summarise the ones we will mention here in Table  \ref{tab:fair_definitions}. Traditionally, the proposed notions have been classified into two categories. The simplest and most well-studied, group fairness, is based on defining distinct protected groups in the given data. Then, for each of these groups, a user-selected statistical constraint must be satisfied. This has notable disadvantages: It requires groups to be treated fairly in aggregate, but this guarantee does not necessarily extend to individuals \cite{Awasthi20}. Further, different statistical constraints prioritise different aspects of fairness. Many of them have also been shown to be incompatible with each other, making the choice even more difficult for users. Finally, the choice of the protected groups that should be considered is an open question \cite{Blum18, Kleinberg17}.

An orthogonal notion to group fairness is individual fairness. Put simply, this notion requires that "similar individuals be treated similarly" \cite{Dwork12}. This approach addresses the previous lack of any individual-level guarantees. However, it requires strong functional assumptions and still requires the step of choosing an underlying metric over the dataset features \cite{Awasthi20}. 

Alternative models of fairness have been proposed to address the disadvantages of the two traditional definitions. One model is causal fairness, which examines the unfair causal effect the sensitive attribute value may have on the prediction made by an algorithm \cite{Mhasawade21}. Another, which is explored in this paper, is differential fairness (DF). This is an extension of the established group fairness concepts that applies them to the case of intersectionalities, meaning groups that are defined by multiple overlapping sensitive attributes \cite{Foulds20, Morina19}. 

A similar model is statistical parity subgroup fairness (SF), which focuses on mitigating intersectional bias by applying group fairness to the case of infinitely many, very small subgroups \cite{Kearns18}. SF and DF are notable because they both enable a more nuanced understanding of unfairness than when a single sensitive attribute and broad, coarse groups are considered. A key difference between them, however, is DF's focus on minority groups. The SF measure of subgroup parity weighs larger groups more heavily than very small ones, while DF-parity considers all groups equally. This means DF can provide greater protection to very small minority groups since, in SF, their impact on the overall score is reduced \cite{Foulds20}.

Despite the lack of consensus on any universal notion of fairness, research has proceeded using the existing models. A major line of work in the development of fair learning algorithms is enforcing fairness during the training step of a classifier \cite{Donini18}. A simple yet effective algorithm that follows this strategy is Calders and Verwer's two-naive-Bayes algorithm \cite{Calders10} (2NB). This algorithm was originally proposed as one of three ways of pursuing fairness in naive Bayes classification. It received further attention in the 2013 publication \cite{Kamishima13} which asserted its effectiveness in enforcing group fairness in binary classification and explored its underlying statistics. It works by training separate naive Bayes classifiers for each of the two (by assumption) groups comprise the dataset, the privileged and the non-privileged group. Then, the algorithm iteratively assesses the fairness of the combined model and makes small changes to the observed probabilities in the direction of making them more fair \cite{Friedler19}.

A recent publication exploring the arguments for and against statistical parity \cite{Raz21} has served as motivation to re-visit algorithms based around it. Statistical parity (also referred to as demographic parity or independence) is a group fairness notion which requires that the groups comprising the dataset receive positive labels with the same likelihood. An assumption that is at the core of 2NB and many other research works, however, is that of a single, binary sensitive feature \cite{Oneto20}. This assumption has been noted to rarely hold in the real world, and eliminating it is one of the essential goals of the previously introduced notions of differential fairness and subgroup parity fairness \cite{Foulds20, Kearns18}. 

This opens the question of how 2NB can be applied to data with multiple, overlapping sensitive attributes while avoiding oversimplification. The 2NB algorithm is applicable to a wide range of tasks and its effectiveness, even in comparison to more complex algorithms, has been demonstrated \cite{Kamishima13, Friedler19}. At the same time, its' design is sufficiently elegant and intuitive to be approachable to practitioners across many disciplines - an important advantage. Thus, extending the algorithm to cover more use cases will be the focus of this work. 

{
\renewcommand{\arraystretch}{1.5}
\begin{table}[ht]
\centering
  \begin{tabular}{p{0.2\linewidth}| p{0.79\linewidth}}
    \hline
    Name & Definition\\
    \hline
    Statistical Parity & Likelihood of positive prediction given group membership should be equal for all groups.\\

    Disparate \newline Impact & Mean ratio of positive predictions for each pair of groups should be 1 or greater than $p\%$.\\

    Subgroup Fairness & Group fairness applied to infinite number of very small groups.\\

    Differential \newline Fairness & Group fairness applied to groups defined by multiple overlapping sensitive attributes. \\

    Individual Fairness & Distance between the likelihood of outcomes between any two individuals should be no greater than similarity distance between them.\\
  
    Causal \newline Fairness & Use of causal modelling to find effect of sensitive attributes on predictions.\\
    \hline
\end{tabular}
  \caption{Some notable formalisations of fairness.}
  \label{tab:fair_definitions}
\end{table}
}

\subsection{Contributions}
This paper seeks to build upon Calders and Verwer's work by exploring the following:
\begin{itemize}
    \item We adapt the original 2NB structure and balancing routine to support multiple, polyvalent (categorical) sensitive features.
    \item We use this new property of the algorithm to apply it to differential fairness.
    \item To support the above, we examine the extended algorithm's performance on real-world US Census data.
    \item Finally, we lay out important considerations users should be aware of before using this algorithm. We draw upon the literature to lay out the pros, cons, and ethical implications of using statistical parity as a fairness criterion.
\end{itemize}

\subsection{Related Work}
\paragraph{Naive Bayes} Naive Bayes is a probabilistic data mining and classification algorithm. In spite of its relative simplicity, it has been shown to be very competent in real-world applications that require classification or class probability estimation and ranking\footnote{Recent, novel applications include \cite{NB_Diaz19, NB_Feng2018, NB_Niazi19} among others.}.
Various strategies have been explored for improving the algorithm's performance by weakening its conditional independence assumption. These include structure extension, attribute weighting, etc. These techniques focus on maximising accuracy or averaged conditional log likelihood \cite{Jiang11}.
Calders and Verwer's proposal of composing multiple naive Bayes models instead aims to enforce independence of predictions with respect to a binary sensitive feature, thus satisfying the statistical parity constraint between the two groups \cite{Calders10}.

\paragraph{Fair Classification} There is a large body of research into designing learning methods that do not use sensitive information in discriminatory ways \cite{Oneto20}. As mentioned, various formalisations of fairness exist but the most well-studied one is group fairness \cite{Blum18}. Many algorithms designed around this notion are introduced as part of the comparative experiment in Section \ref{section:exp}.

A more recent proposal, differential fairness (DF), extends existing group fairness concepts to protect subgroups defined by intersections of and by individual sensitive attributes. The original papers by \cite{Foulds20} and \cite{Morina19} explore the context of intersectionality, and provide comparisons of DF with established concepts. The first paper asserts DF's distinction from subgroup parity and demonstrates its usefulness in protecting small minority groups. The latter paper gives methods to robustly estimate the DF metrics and proposes a post-processing technique to enforce DF on classifiers. 

\paragraph{Humanistic Analysis} A line of work that is parallel to fair algorithm development focuses on analysing these proposals from an ethical, philosophical, and moral standpoint. A recent such publication, which examines statistical parity among other notions, and which motivated and influenced this paper, is by Hertweck, Heitz, and Loi \cite{Hertweck21}. They propose philosophically-grounded criteria for justifying the enforcement of independence/statistical parity in a given task. They include scenarios where enforcing statistical parity is ethical and justified, as well as counter-examples where the criteria are met but independence should not be enforced. As with many similar works, they conclude by directing the reader to strike a balance between fairness and utilitarian concerns (such as accuracy) in their task.
\cite{Heidari19} do similar work, laying out the moral assumptions underlying several popular notions of fairness. In \cite{Raz21}, R\"az critically examines the advantages and shortcomings of statistical parity as a fairness criterion and makes an overall positive case for it.

\cite{Frielder16} introduce the concept of distinct \emph{worldviews} which influence how we pursue fairness. One of them is that We're All Equal (WAE) i.e. there is no association between the \emph{construct} (the latent feature that is truly relevant for the prediction) and the sensitive attribute. The orthogonal worldview is that \emph{What You See Is What You Get}, wherein the observed labels are accurate reflections of the construct.  In \cite{Yeom21}, Yeom and Tschantz give a measure of \emph{disparity amplification} and dissect the popular group fairness models of statistical parity, equalised odds, calibration, and predictive parity through the lens of worldviews. They argue that under WAE, statistical parity is required to eliminate disparity amplification. However, deviating from this worldviews introduces inaccuracy when we enforce parity.

\section{N-Naive-Bayes Algorithm}
The proposed N-naive-Bayes algorithm is a supervised binary classifier that allows the enforcement of a statistical fairness constraint in its predictions. Given an (ideally large) training set of labelled instances, the algorithm partitions the data based on sensitive attribute value and trains a separate naive Bayes sub-estimator on each of the sub-sets. This is an extension of the original two-naive-Bayes structure, where exactly two sub-estimators are trained. 
The next step of the training stage is for the conditional probabilities $P(Y|S)$ to be empirically estimated from the training set. Where $N_s$ is the number of instances that belong to group $s$, and $N_{y,s}$ the number of instances of that group that have label $y$, the empirical conditional probability\footnote{Equation (\ref{eqn:pys}) gives a smoothed empirical probability, where the constant $\alpha$ is the parameter of a symmetric Dirichlet prior with concentration parameter $2*\alpha$, since a binary label is assumed.} is given as:

\begin{equation}
    \label{eqn:pys}
    P(y|s) = \frac{N_{y,s} + \alpha}{N_s + 2*\alpha}
\end{equation}

Finally, the algorithm modifies the joint distribution $P(Y,S)$ to enforce the given fairness constraint. Then, the final predicted class probabilities, for a sample $xs$ (where $x$ is the feature vector excluding the sensitive feature $s$), is:

\begin{align}
    \label{eqn:cxs}
    P(y|xs) &= P(x|y) * P(s|y) * P(y)\\
    &= C_s(x) * P(s|y) * P(y)\\
    &= C_s(x) * P(s \cap y)
\end{align}
Where $C_s$ is the the sub-estimator for sensitive group $s \in S$.

\subsection{Enforcing Statistical Parity} To satisfy the statistical parity constraint, the original 2NB algorithm runs a heuristic post-processing routine that iteratively adjusts the conditional probabilities $P(Y|S)$ of the groups in the direction of making them equal. During its execution, this probability-balancing routine alternates between reducing $N(Y=1,S=1)$ and increasing $N(Y=1,S=0)$ depending on the number of positive labels outputted by the model at each iteration. This is to try and keep the resultant marginal distribution of $Y$ stable. Once balancing is complete, the value of $P(S|Y)$ can be induced from $N_{y,s}$ similar to (\ref{eqn:pys}). The first contribution of this paper is to extend this routine to suit the polyvalent definition of statistical parity we will use:

\begin{definition} Statistical (Conditional) Parity for Polyvalent $S$ \cite{Ritov17}: 

For predicted binary labels $\hat{y}$ and polyvalent sensitive feature $S$, statistical (conditional) parity requires \footnote{The cited definition requires this to hold for all values of $\hat{y}$, however for a binary label it is sufficient to check $\hat{y} = 1$.}:
\begin{equation}
P(\hat{y}=1|s) = P(\hat{y}=1|s') \; \forall \; s, s' \in S
\end{equation}
\end{definition}

We modify the probability-balancing routine to subtract and add probability to the group with the highest (max) and lowest (min) current $P(Y=1|s)$ respectively. These probabilities are re-computed with each iteration, and the max and min groups re-selected. Further, we introduce the constraint that only groups designated by the user as privileged can receive a reduction in their likelihood of getting a positive label \footnote{A similar constraint is explored by \cite{Zafar17}.}. This is to avoid making any assumptions about which groups it would be appropriate to demote positive instances of. It allows the balancing routine to terminate immediately if it over-corrects, or if the data is such that $P(\hat{y}=1|s_{np}) > P(\hat{y}=1|s_p)$ to begin with, as is the case in the well-known UCI Adult dataset, for example. This gives us the final form of our statistical parity criterion:

\begin{definition}
Statistical Parity Criterion for NNB:

\label{eqn:discscore}
For predicted binary labels $\hat{y}$ and sensitive feature $S$:
\begin{equation}
    P(\hat{y}=1|s_p) = P(\hat{y}=1|s_{np}) \; \forall \; (s_p, s_{np}) \in S_p \times S_{np}
\end{equation} 
Where $S_p$ and $S_{np}$ are the sub-sets of all privileged and non-privileged sub-groups of $S$ respectively. 
\end{definition}

We adapt the above definition into a score that the algorithm can minimise: 

\begin{equation}
    disc = \max P(\hat{y}=1|s_p) - \min P(\hat{y}=1|s_{np})
    \label{eqn:criterion}
\end{equation}

Note that the above criterion can easily be relaxed to apply the four-fifths rule for removing disparate impact (or its more general form, the $p\%$ rule \cite{Zafar17}) instead of perfect statistical parity. For the purposes of this paper, however, we explore the effect of statistical parity in its base form.

We also note the definition of disparate impact we use in the evaluation stage:
\begin{definition} Disparate Impact (Mean) for Polyvalent $S$:
\begin{equation*}
    \frac{1}{|S_p \times S_{np}|} \sum_{(s_p, s_{np})} \frac{P(\hat{y} = 1|s_{np})}{P(\hat{y}=1|s_p)}
\end{equation*}
\end{definition}

Algorithm \ref{alg:parity} describes the extended probability balancing heuristic for enforcing parity. The values of $s_p, s_{np}$ in the parity criterion (Equation \ref{eqn:criterion}) are referred to as $s_{max}$ and $s_{min}$ respectively. At each iteration, the routine determines these groups and adjusts their conditional probabilities. A further modification from the original is that the proportion by which the probabilities are adjusted with each iteration is now proportional to the size of the group itself, instead of the size of the opposite group. In experiments, this yields a great performance improvement, especially where the distribution of samples over $S$ is very imbalanced. 

\begin{algorithm}[tb]
    \caption{Pseudocode for a probability-balancing routine to enforce statistical parity}
    \label{alg:parity}
    \begin{algorithmic}[1]
        \STATE Calculate the parity score, $disc$, of the predicted classes by the current model and store $s_{max}$, $s_{min}$
        \WHILE{$disc > disc_0$}
            \STATE Let $numpos$ be the number of positive samples by the current model
            \IF {$numpos < \;$ the number of positive samples in the training set}
                \STATE $N(y=1, s_{min}) \; += \; \Delta * N(y=0, s_{min})$
                \STATE $N(y=0, s_{min}) \; -= \; \Delta * N(y=0, s_{min})$
            \ELSE
                \STATE $N(y=1, s_{max}) \; += \; \Delta * N(y=1, s_{max})$
                \STATE $N(y=0, s_{max}) \; -= \; \Delta * N(y=1, s_{max})$
            \ENDIF
            \STATE If any $N(y,s)$ is now negative, rollback the changes and terminate
            \STATE Recalculate $P(Y|S)$, $disc$, $s_{max}$, $s_{min}$
        \ENDWHILE
    \end{algorithmic}
\end{algorithm}

\subsection{Enforcing Differential Fairness}
\label{section:df}

An alternative measure of fairness we explore is differential fairness, as given in \cite{Foulds20}. 

\begin{definition}
    \label{def:epsilon}
    A classifier is $\epsilon$-differentially fair if:
    
    \begin{equation}
        e^{-\epsilon} \leq \frac{P(\hat{y}|s)}{P(\hat{y}|s')} \leq e^{\epsilon} \; \forall \; s,s' \in S, \hat{y} \in Y
    \end{equation}
\end{definition}

The (smoothed) empirical differential fairness score, from the empirical counts in the data, assuming a binary label, is:

\begin{equation}
    e^{-\epsilon} \leq \frac{N(\hat{y}, s) + \alpha}{N(s) + \beta} \frac{N(s') + \beta}{N(\hat{y}, s') + \alpha} \leq e^{\epsilon} \; \forall \; s, s' \in S, \hat{y} \in Y
\end{equation}

This is used in experiments to estimate the value of $\epsilon$ (the $\epsilon$-score) from the predicted labels on the dataset\footnote{Note that this definition produces noisier estimates for subgroups with fewer members. \cite{Morina19} shows that as the dataset grows, the given estimate converges to the true value, and that this happens regardless of the chosen smoothing parameters. However, for small or imbalanced datasets, more robust estimation methods should be used. }. In experiments we set $\beta = 2*\alpha$ and substitute with the observed conditional probability estimates from the dataset.
An additional measure given in \cite{Foulds20} to assess fairness from the standpoint of intersectionality is \emph{differential fairness bias amplification}. This measure gives an indication of how much a black-box classifier increases the unfairness over the original data \cite{Foulds20,Zhao17}.

\begin{definition}
    \label{def:biasamp}
    Differential Fairness Bias Amplification
    
    A classifier C satisfies $(\epsilon_2-\epsilon_1)$-DF bias amplification w.r.t. dataset $D$ if $C$ is $\epsilon_2$-DF fair and $D$ is $\epsilon_1$-DF fair. 
\end{definition}

To adjust the joint distribution $P(Y,S)$ to minimise satisfy DF-fairness and minimise the $\epsilon$-score, we propose a new heuristic probability-balancing routine and associated discrimination score. The distinction from the balancing routine given in Algorithm \ref{alg:parity} is that this focuses on outputting a narrower range of probabilities, while still avoiding negatively impacting groups that are designated as non-privileged. To form the new discrimination score, we apply the principle of separating privileged and non-privileged sub-groups of S from the previous section to the $\epsilon$-score definition:

\begin{equation}
    e^{-\epsilon} \leq \frac{P(\hat{y}=1|s_{np})}{P(\hat{y}=1|s_p)} \leq e^{\epsilon} \; \forall \; (s_p, s_{np}) \in S_p \times S_{np}
\end{equation}

We then express this restricted $\epsilon$-score as the maximum of two ratios: $e^{\epsilon} = max(\rho_d, \rho_u)$, where for $(s_p, s_{np}) \in S_p \times S_{np}$:

\begin{equation}
    \rho_d = \max\frac{P(\hat{y}=1|s_{np})}{P(\hat{y}=1|s_{p})}, \; \rho_u = \max\frac{P(\hat{y}=1|s_{p})}{P(\hat{y}=1|s_{np})} 
\end{equation}

The execution of the proposed balancing routine is determined by these ratios. If $\rho_d$ is greater, then the non-privileged sub-group with smallest probability at that iteration receives an increase in probability. If $\rho_u$ is greater, then the privileged group with highest probability receives a decrease in probability. These conditions can be expected to alternate as the conditional probabilities $P(Y|S)$ converge. Iteration continues until $\rho_d$ is close to zero. The $s_{max}$ and $s_{min}$ groups are determined as in the previous section.

\begin{algorithm}[tb]
    \caption{Pseudocode for a probability-balancing routine to enforce DF parity}
    \label{alg:df}
    \begin{algorithmic}[1]
        \STATE Calculate the ratios $\rho_d, \rho_u$ empirically from the predicted classes by the current model, store $s_{max}$, $s_{min}$
        
        \WHILE{$\rho_d > disc_0$}
            \IF {$\rho_u \leq \rho_d$}
                \STATE $N(y=0, s_{min}) \; -= \; \Delta * N(y=0, s_{min})$
                \STATE $N(y=1, s_{min}) \; += \; \Delta * N(y=1, s_{min})$
            \ELSE
                \STATE $N(y=0, s_{max}) \; += \; \Delta * N(y=0, s_{max})$
                \STATE $N(y=1, s_{max}) \; -= \; \Delta * N(y=1, s_{max})$
            \ENDIF
            \STATE Recalculate $P(Y|S)$, $\rho_d$, $\rho_u$, $s_{max}$, $s_{min}$
        \ENDWHILE
    \end{algorithmic}
\end{algorithm}

This routine disregards the number of positive labels the model produces, while Algorithm \ref{alg:parity} attempts to keep that number close to the number of positive labels in the training data. This allows it to avoid situations where a single, non-privileged sub-group with small probability would require the probabilities of the privileged groups to be reduced significantly. In such cases, other non-privileged sub-groups might maintain much higher probabilities, therefore giving a poor $\epsilon$-score. An further difference is the proportion by which each $N_{y,s}$ is modified grows/decreases exponentially. In experiments, this allows the routine escape local minima that occur during the adjustment of $P(Y|S)$ and lead to inefficiency. This routine does, however, offer a theoretical accuracy trade-off compared to Algorithm \ref{alg:parity}, which we investigate in the following section.

Finally, note that all the above probability-balancing routines (including Calders and Verwer's original one) are based around the assumption that the distribution of labels over the sensitive feature(s) in the training set is reflective of the test setting. This assumption is not unique to this model (see \cite{Agarwal18, Hardt16}), and under it, we can conclude that minimising the given fairness measure on the training set generalises to the test data \cite{Singh21}.

\section{Experimental Results}
\label{section:exp}
\subsection{Setup}
We implement NNB in Python within the scikit-Learn framework, using Gaussian naive Bayes as the sub-estimator. We then evaluate its performance in two experiments.

For both experiments, we use real-world data from the US Census Bureau\footnote{https://www.census.gov/programs-surveys/acs/microdata/documentation.html}. \cite{Ding21} define several classification tasks on this data, each involving a sub-set of the total features available. We consider two:
\begin{itemize}
    \item \verb|Income|: Predict whether an individual's income is above \$50,000. The data for this problem is filtered so that it serves as a comparable replacement to the well-known UCI Adult dataset.
    \item \verb|Employment|: Predict whether an individual is employed
\end{itemize}

The details of which features are included in each task and what filtering takes place can be found in the paper \cite{Ding21} and the associated page on GitHub\footnote{https://github.com/zykls/folktables}. To evaluate NNB we use data from the 2018 census in the state of California. The sensitive feature(s) used in each task are indicated after its name, e.g. \verb|Income-Race-Sex| is the \verb|Income| task using race and sex as the sensitive features. To best capture intersectional fairness when using multiple sensitive features, we follow the approach from \cite{Foulds20} and define each group $s$ as a tuple of the sub-groups of each sensitive feature that each sample belongs to.

\paragraph{First Experiment} This experiment compares NNB's performance with other algorithms. The comparison includes "vanilla" models as baselines for performance, and several group-fairness-aware algorithms that have a similar focus to NNB - ensuring non-discrimination across protected groups by optimising metrics such as statistical parity or disparate impact. Specifically, we consider the following:
\begin{itemize}
    \item \verb|GaussianNB|, \verb|DecisionTree|, \verb|LR|, \verb|SVM|: scikit-Learn's Gaussian naive Bayes, Decision Trees, Logistic Regression, and SVM.
    \item \verb|Feldman-DT|, \verb|Feldman-NB|: A pre-processing algorithm that aims to remove disparate impact. It equalises the marginal distributions of the subsets of each attribute with each sensitive value \cite{Feldman15}. The resulting "repaired" data is then used to train scikit-Learn classifiers - Decision Trees (\verb|DT|) and Gaussian naive Bayes (\verb|NB|).
    \item \verb|Kamishima|: An in-processing method that introduces a regularisation term to logistic regression to enforce independence of labels from the sensitive feature \cite{Kamishima12}.
    \item \verb|ZafarAccuracy|, \verb|ZafarFairness|: An in-processing algorithm that applies fairness constraints to convex margin-based classifiers \cite{Zafar17} . Specifically, we test two variations of a modified logistic regression classifier: The first maximises accuracy subject to fairness (disparate impact) constraints, while the latter prioritises removing disparate impact.
    \item \verb|2NB|: Calders and Verwer's original algorithm, using the same GaussianNB sub-estimator as NNB. 
    \item \verb|NNB-Parity|, \verb|NNB-DF|: N-naive-Bayes tuned to satisfy statistical parity using Algorithm \ref{alg:parity}, and DF-parity using Algorithm \ref{alg:df}.
\end{itemize}

For the comparison we use the benchmark provided by \cite{Friedler19}. The fairness-aware algorithms are tuned via grid-search to optimise accuracy. The performance of the algorithms is then measured over ten random train-test splits of the data.
\begin{figure*}[t]
  \centering
  \includegraphics[width=0.9\textwidth]{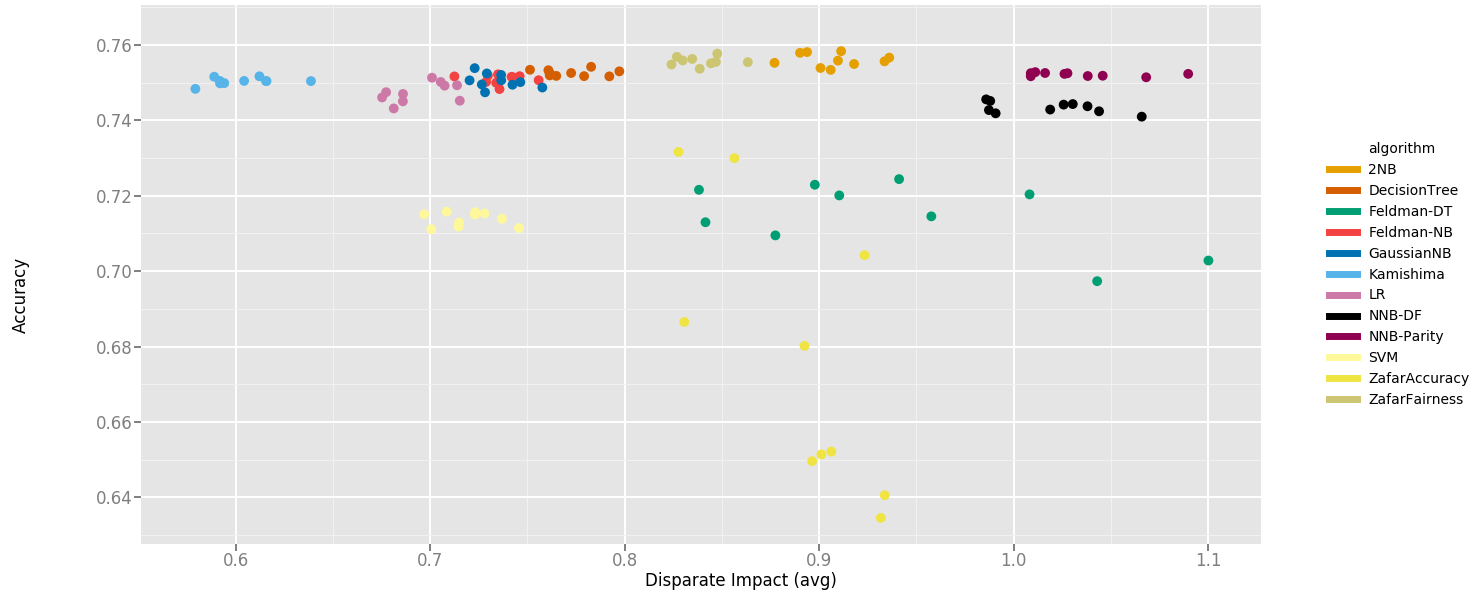}
  \includegraphics[width=0.9\textwidth]{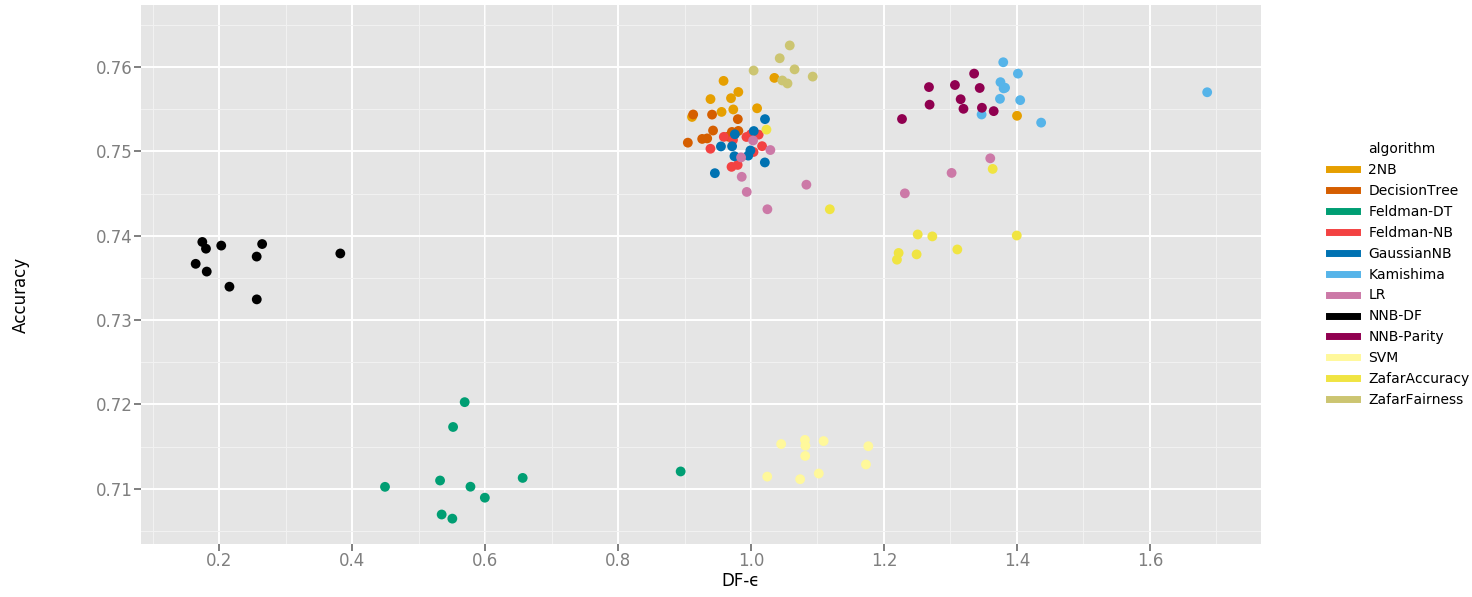}
  \caption{Scatter plots of accuracy vs. disparate impact for Income-Race and vs. $\epsilon$-score for Income-Race-Sex}
    \label{fig:plot1}
\end{figure*}
\paragraph{Second Experiment} This experiment demonstrates how NNB performs in finer detail. We consider \verb|GaussianNB|, \verb|NNB-Parity|, and \verb|NNB-DF| as before, and we further include \verb|2NB|, the original two-naive-Bayes algorithm implemented identically to NNB. Finally, we include \verb|Perfect| as a secondary baseline, to illustrate the scores that would be achieved by a perfect classifier.

To evaluate the performance of the above algorithms, we note the mean and variance of the following measures over 10 random train-test splits: accuracy, AUC, disparate impact score (mean of the DI between all privileged and non-privileged groups), statistical parity score (as defined in \ref{eqn:discscore}), DF-$\epsilon$ (as defined in \ref{def:epsilon}), DF-bias amplification score (as defined in \ref{def:biasamp}). We also compare the resultant distribution of labels over groups of $S$ on a single random train-test split.

\begin{figure*}[t]
  \centering
  \includegraphics[width=0.9\textwidth]{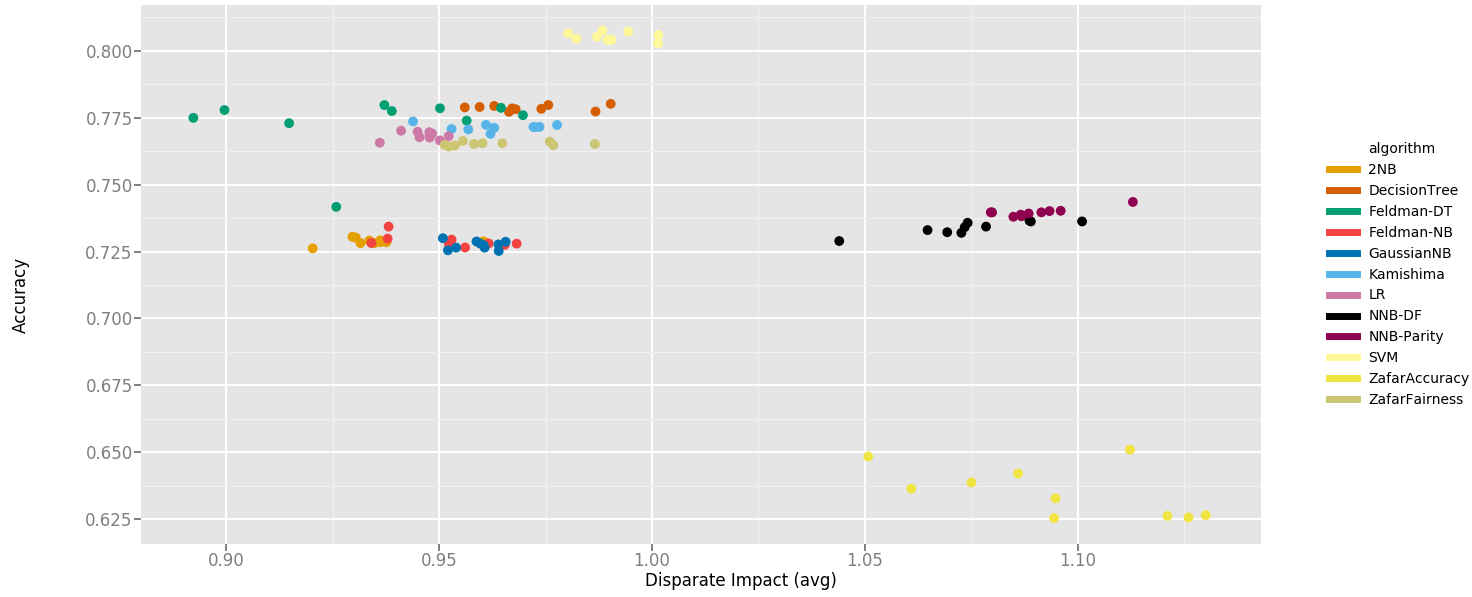}
  \includegraphics[width=0.9\textwidth]{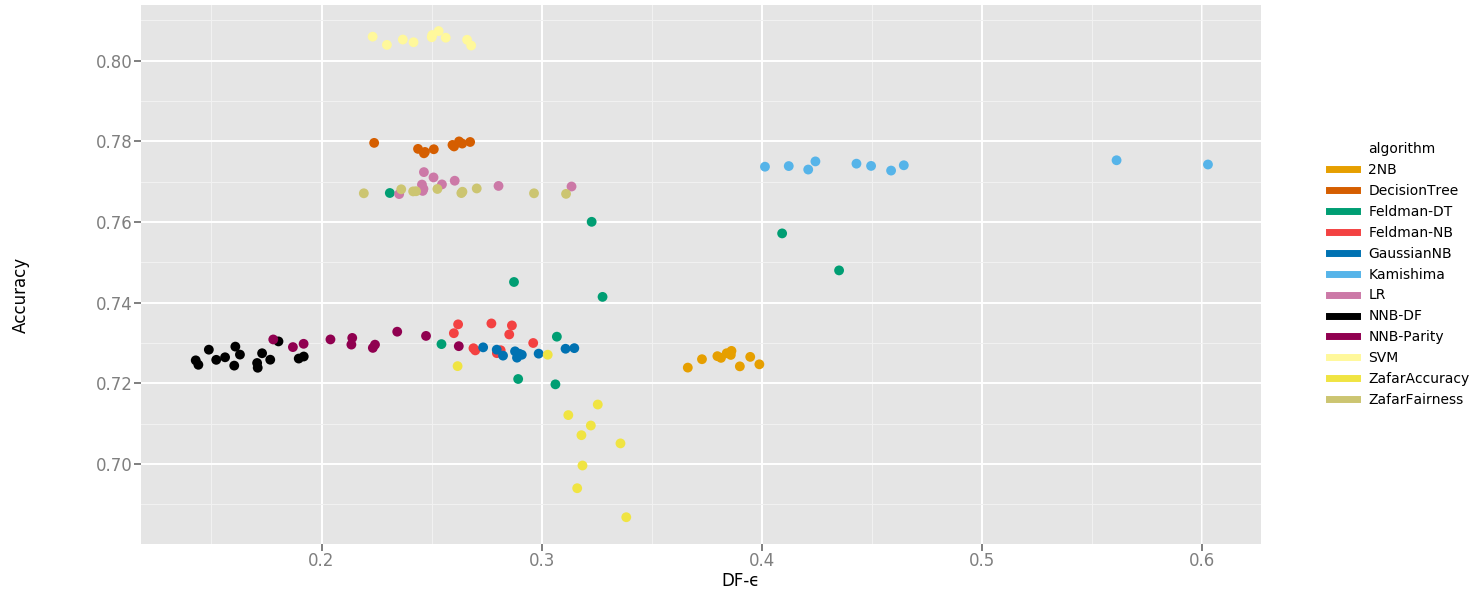}
  \caption{Scatter plots of accuracy vs. disparate impact for Employment-Race and  vs. $\epsilon$-score for Employment-Race-Sex}
    \label{fig:plot2}
\end{figure*}

{
\renewcommand{\arraystretch}{1.2}
\begin{table*}[t]\fontsize{9}{11}
\centering
  \begin{tabular}{l|cccccc}
    \hline
    & AUC & Accuracy & DI & Parity & DF-$\epsilon$ & DF-amp\\
    \hline
    \verb|GaussianNB| & $0.8270 \pm 0.00$ & $0.7503 \pm 0.00$ & $0.6304 \pm 0.0001$ & $0.4222 \pm 0.0000$ & $1.4100 \pm 0.0012$ & $0.4680 \pm 0.0045$\\
    \verb|2NB| & $0.8223 \pm 0.00$ & $0.7577 \pm 0.00$ & $0.8930 \pm 0.0013$ & $0.3606 \pm 0.0000$ & $0.9774 \pm 0.0016$ & $0.0353 \pm 0.0059$\\
    \verb|NNB-Parity| & $0.8114 \pm 0.00$ & $0.7480 \pm 0.00$ & $1.0810 \pm 0.0006$ & $0.1984 \pm 0.0005$ & $0.4580 \pm 0.0045$ & $-0.4840 \pm 0.0041$\\
    \verb|NNB-DF| & $0.8138 \pm 0.00$ & $0.7380 \pm 0.00$ & $1.0636 \pm 0.0007$ & $0.1530 \pm 0.0008$ & $0.3112 \pm 0.0048$ & $-0.6308 \pm 0.0035$\\
    \verb|Perfect| & $1.0000 \pm 0.00$ & $1.0000 \pm 0.00$ & $0.6975 \pm 0.0005$ & $0.2950 \pm 0.0001$ & $0.9420 \pm 0.0048$ & $0.0000 \pm 0.0000$\\
  \hline
\end{tabular}
  \caption{Scores Achieved on Income with Race as the Sensitive Feature}
  \label{tab:scores1}
\end{table*}

\begin{table*}[t]\fontsize{9}{11}
\centering
 
  \begin{tabular}{l|cccccc}
    \hline
    & AUC & Accuracy & DI & Parity & DF-$\epsilon$ & DF-amp\\
    \hline
    \verb|GaussianNB| & $0.8159 \pm 0.00$ & $0.7273 \pm 0.00$ & $1.0228 \pm 0.0001$ & $0.3000 \pm 0.0001$ & $0.4994 \pm 0.0003$ & $0.0922 \pm 0.0016$\\
    \verb|2NB| & $0.8112 \pm 0.00$ & $0.7202 \pm 0.00$ & $0.9352 \pm 0.0001$ & $0.2951 \pm 0.0001$ & $0.4818 \pm 0.0002$ & $0.0746 \pm 0.0015$\\
    \verb|NNB-Parity| & $0.7820 \pm 0.00$ & $0.7241 \pm 0.00$ & $1.2990 \pm 0.0004$ & $0.2478 \pm 0.0007$ & $0.3971 \pm 0.0013$ & $-0.0101 \pm 0.0005$\\
    \verb|NNB-DF| & $0.7909 \pm 0.00$ & $0.7251 \pm 0.00$ & $1.0601 \pm 0.0002$ & $0.1272 \pm 0.0009$ & $0.1840 \pm 0.0018$ & $-0.2232 \pm 0.0011$\\
    \verb|Perfect| & $1.0000 \pm 0.00$ & $1.0000 \pm 0.00$ & $0.8643 \pm 0.0001$ & $0.1782 \pm 0.0002$ & $0.4072 \pm 0.0014$ & $0.0000 \pm 0.0000$\\
  \hline
\end{tabular}
  \caption{Scores Achieved on Employment with Race and Sex as the Sensitive Features}
  \label{tab:scores2}
\end{table*}
}

\subsection{Results}

\paragraph{First Experiment} 
Figure \ref{fig:plot1} gives the accuracy vs. the disparate impact and DF-$\epsilon$ scores on the Income-Race and Income-Race-Sex tasks. Figure \ref{fig:plot2} shows the same for Employment-Race and Employment-Race-Sex. It can be seen that on Income-Race, NNB results in a higher DI score than 2NB and has often over-favoured non-privileged groups causing a score $>1$. Its accuracy is on-par with 2NB and the baseline naive Bayes, DT, and LR models. Feldman's algorithm with Decision Trees results similar disparate impact score in some splits, but lower accuracy. The same is true for the DF-$\epsilon$ score on this task. On Income-Race-Sex, \verb|NNB-DF| beats out all other algorithms in achieving DI $\sim 1$, however \verb|NNB-Parity| has higher accuracy than both \verb|NNB-DF| and naive Bayes. \verb|NNB-DF| is also the most successful at minimising the $\epsilon$-score for this task, though again this comes at the cost of lower accuracy than the baseline model.

On Employment-Race all naive Bayes models achieve similar accuracy, while DT and LR-based models rank higher, and SVM the highest. The same can be observed for Employment-Race-Sex, and for both tasks \verb|NNB-DF| again gives the $\epsilon$-scores closest to zero. 

\paragraph{Second Experiment}

Table \ref{tab:scores1} gives the scores achieved on the \verb|Income-Race| task, and Table \ref{tab:scores2} gives the same \verb|Employment-Race-Sex|. On \verb|Income-Race|, both NNB models gave an improved parity score compared to the perfect classifier and \verb|GaussianNB|. NNB and 2NB also gave improved disparate impact scores over the baseline models, but 2NB under-corrected while the NNB models gave a score $>1$ indicating they favoured the non-privileged groups over the privileged group.

\verb|NNB-Parity| and \verb|NNB-DF| gave similar disparate impact scores, but the former gave higher accuracy while the latter produced a narrower range of positive label proportions, and thus better parity, $\epsilon$, and DF-bias amplification scores. The evident accuracy trade-off is more pronounced in the latter task, with \verb|NNB-Parity| achieving an accuracy of $0.7445 \pm 0.00$, and \verb|NNB-DF| achieving $0.7199 \pm 0.00$.

On \verb|Employment-Race-Sex|, \verb|NNB-DF| outperformed \verb|NNB-Parity| on all scores. This was also the case for \verb|Employment-Race|, where both models had similar accuracy but \verb|NNB-DF| displayed less over-correction in its disparate impact score ($1.0336 \pm 0.0001$ versus $1.2760 \pm 0.0002$), in addition to the expected improvement in $\epsilon$-score ($0.1068 \pm 0.001$ versus $0.3434 \pm 0.0001$). This suggests the DF balancing routine is better suited for the \verb|Employment| task than the parity-based routine. 

\section{Discussion}
In this work we presented an extension of the two-naive-Bayes algorithm, adapting it to suit datasets with multiple, polyvalent sensitive features. We applied the proposed N-naive-Bayes structure to intersectionality and differential fairness by giving an alternative probability-balancing routine. Our experiments on real-world datasets yielded favourable results and demonstrated the effectiveness and the differences between the parity and DF-based approaches.

We conclude by laying out key considerations users should take into account before using N-naive-Bayes:

\paragraph{Statistical Parity as a Fairness Criterion} Statistical parity stands opposed to the (aggregate) accuracy of a classifier, except in degenerate cases where the data is already fair, so it is recommended that a balance between the two is pursued \cite{Hertweck21}. This also applies to the extended, but still parity-based, DF measure that was explored in Section \ref{section:df}. In their worldview-based analysis, Yeom and Tschantz caution us that even under WAE, blind enforcement of statistical parity can introduce new discrimination into the system \cite{Yeom21}. Thus, users must be aware of the ethical implications of using parity as a core fairness constraint, the possible impact it may have on individuals, and the moral objections these individuals may justifiably raise. 

We recommend further reading on the advantages and disadvantages of group fairness in general \cite{Raz21, Dwork12, Heidari19}, as well as parity specifically \cite{Hertweck21, Yeom21}, so users can make informed decisions on how to apply statistical parity and N-naive-Bayes to their application.

\paragraph{Limitations of NNB} N-naive-Bayes (as with two-naive-Bayes) has inherent limitations. The algorithm does not automatically make a classification task fair when it is applied. This is only considered to be possible by doing extensive domain-specific investigation \cite{Hardt16}. Rather, the algorithm introduces a form of affirmative action to the task, increasing and decreasing the likelihood of different groups receiving a positive label in an attempt to satisfy the given parity constraint. This intentional manipulation of the original distribution over the data can be done to correct for structural biases in the data, for the purposes of compliance with regulations, or even as part of an effort to counteract historical inequalities.

Users should always consider the implications of estimating probability distributions for each group separately (as is done at the beginning of the training stage), as well as the mechanism behind any post-facto probability tuning they decide on. Further, users should understand the implications of affirmative action, its downstream effects, and ensure it is appropriate to their application. As a starting point for further reading, see \cite{Dwork12, Kannan19}. Sociological and legal works such as \cite{Kalev06, Anderson03} are also recommended.

Finally, the explicit choice of sensitive features to consider when enforcing statistical parity is a simplification of the real world and should be done carefully. One should consider the ontology behind observed values in the dataset: race, for example, has varying definitions, each of which comes with its own assumptions. Further, identifying groups in the data using a set of observable qualities, whatever those may be, also carries implicit assumptions about how all the factors involved interact with each other and the validity of decomposing them into discrete features \cite[Ch.~5]{barocas-hardt-narayanan}.

\bibliography{references.bib}
\end{document}